\documentclass[
twocolumn,
]{ceurart}
\usepackage{graphicx}
\usepackage{amsmath}
\usepackage{amssymb}
\usepackage{amsthm}
\usepackage{booktabs}
\usepackage{algorithm}
\usepackage{algpseudocode}
\usepackage{multirow}
\usepackage{makecell}
\usepackage{longtable}
\usepackage{arydshln}
\usepackage{xcolor}
\usepackage{float}
\usepackage{xspace}
\usepackage{bm, subfigure}

\begin{document}

\copyrightyear{2021}
\copyrightclause{Copyright for this paper by its authors.
  Use permitted under Creative Commons License Attribution 4.0
  International (CC BY 4.0).}

\conference{International Workshop on Safety \& Security of Deep Learning, IJCAI 2021}

\title{Exploring the Asynchronous of the Frequency Spectra of GAN-generated Facial Images}

\author[1]{Binh M. Le}[%
orcid=0000-0002-4344-3421,
email=bmle@g.skku.edu,
]
\address[1]{Department of Software, Sungkyunkwan University, South Korea}

\author[2]{Simon S. Woo}[%
orcid=0000-0002-8983-1542,
email=swoo@g.skku.edu,
]
\address[2]{Department of Applied Data Science, Sungkyunkwan University, South Korea}

\begin{abstract}
The rapid progression of Generative Adversarial Networks (GANs) has raised a concern of their misuses for malicious purposes, especially in creating fake face images. Although many proposed methods succeed in detecting GAN-based synthetic images, they are still limited by the need for large quantities of the training fake image dataset, and challenges for the detector's generalizability to unknown facial images. In this paper, we propose a new approach that explores the asynchronous frequency spectra of color channels, which is simple but effective for training both unsupervised and supervised learning models to distinguish GAN-based synthetic images. We further investigate the transferability of a training model that learns from our suggested features in one source domain and validates on another target domains with prior knowledge of the features' distribution. Our experimental results show that the discrepancy of spectra in the frequency domain is a practical artifact to effectively detect various types of GAN-based generated images.
\end{abstract}


\maketitle

\section{Introduction}
\begin{figure}[!t]
\centering
\includegraphics[width=2.6in]{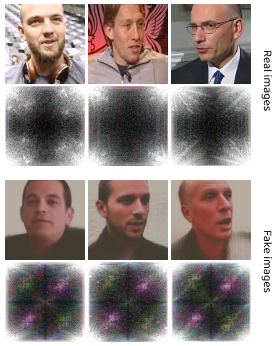}
\caption{The frequency spectra differences of real vs. fake images. \textbf{Top}: Real images and their corresponding concurrent spectra. \textbf{Bottom}: Fake images and their corresponding (chaotic) spectra. It is difficult for human eyes to distinguish between real and fake images, but after applying DFT on each channel of images, the vital clues to distinguish real vs. fake images can be discovered.}
\label{fig:asynchronous_spec}
\end{figure}

In recent years, there has been tremendous progress in Generative Adversarial Networks (GAN), in which two modules (generator vs. discriminator), play a minimax game to produce highly realistic data. Unfortunately, in addition to several fruitful GAN applications, attackers can exploit GANs for malicious purposes, such as spreading fake news \cite{quandt2019fake} or propagating fake pornography of celebrities~\cite{cole_2018} as shown in the past. Meanwhile, several efforts have been made by researchers~\cite{zhang2019detecting,wang2020cnn,frank2020leveraging} to resist these nasty misuses. Wang \emph{et al.} built a deep neural network to classify GAN-based generated images and empirically demonstrates that a classifier trained on one single dataset can generalize to different GAN datasets\cite{wang2020cnn}. Especially, Dzanic and Shah \cite{dzanic2019fourier} empirically show the systematic bias in high spatial frequencies and use this characteristic to classify real and deep network generated images. However, they did not explore the deeper statistical frequency features that we propose in our work, and they simply focused on converting the RGB to gray image, unlike ours.

Also, the checkerboard artifacts in spectrum generated by up-sampling components of GAN model were also extensively investigated by Zhang \emph{et al.} \cite{zhang2019detecting} and Frank \emph{et al.} \cite{frank2020leveraging}. While these techniques have shown success in terms of achieving high accuracy, they typically require a large quantities of data to train and the incurred high computational complexity, which can be prohibitively expensive in many practical applications. As a consequence, there is a strong need for detection methodologies that can achieve comparable levels of performance with limited training data, requiring low computational requirements and better generalization to unknown GAN-based generated images.

In this work, we first observe that the frequencies of the channels in the real images are highly correlated as shown in Fig.~\ref{fig:asynchronous_spec} and Fig.~\ref{fig:his_image_4}. 
In fact, the discriminator in the GAN model can make the generated synthesized images highly realistic, close to real images. However, to the best of our knowledge, there has not been any attempt to apply \textit{direct correlation constraint} between channels on the output images in the frequency domain in the GAN models. From this observation, we hypothesize that the insufficiency of channel-dependent training in most of the current GAN models can produce the channel-wise asynchrony in the frequency domain. The asynchronous can be exposed in various GAN datasets, which can be effectively used to distinguish GAN-based synthetic images by both unsupervised or supervised learning methods. 

In order to demonstrate the effectiveness of our proposed approach, we experiment with four types of datasets: Fake Head Talker \cite{zakharov2019few}, StyleGAN \cite{karras2019style}, StarGAN \cite{choi2018stargan}, and Adversarial Latent Auto Encoder (ALAE) \cite{pidhorskyi2020adversarial}. Our main contributions in this work are summarized as follows: 
\begin{itemize}
\item We firstly introduce the asynchronous in the frequency domain for detecting GAN-generated fake images.
\item We propose effective unsupervised and supervised learning models using our discriminative mining features to classify real and fake images.
\item We demonstrate the transferability of our proposed learning models across different GAN-based synthetic datasets using our spectra features.
\end{itemize}

\section{Our Methods} 
\label{methodology}
\begin{figure*}[t!]
\centering
\includegraphics[width=0.99\linewidth]{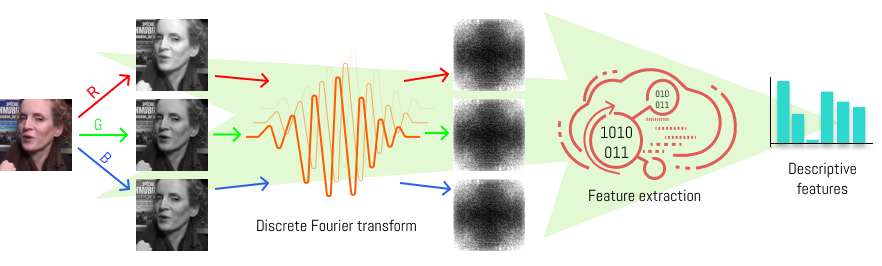}
\caption{Illustration of our end-to-end pipeline. Unlike other research  \protect\cite{dzanic2019fourier} that apply DFT on a grayscale image, we focus on each channel information and utilize statistical methods to obtain discriminative features.}
\label{fig:overall_arc}
\end{figure*}
\subsection{Fourier Spectrum Analysis }
In this section, we construct our hypothesis on the channel-wise asynchrony of GAN-based generated images. The convolutional operation at the $l^{th}$ layer of a generative model is formulated as follows:
\begin{equation}
\label{eqn:fin_conv}
    A_i^{l+1}=Conv_{i}^{l} (A^{l}) = \sigma \left ( \sum_{c=1}^{C_{(l)}} F_{ic}^{l} \circledast Up(A_{c}^{l}) \right),
\end{equation}
where $C_{(l)}$ is the number of channels of the $l^{th}$ layer's output $A^{l}$, $F^{l} \in \mathbb{R}^{k_{l} \times k_{l} \times C_{(l+1)} \times C_{(l)}}$ is a set of $C_{(l+1)} \times C_{(l)}$ trainable $2D$ filters that have size of $k_{l} \times k_{l}$. And $Up(\cdot)$, $\circledast$ and $\sigma(\cdot)$ denote the up-sampling operator, convolutional operator and activation function, respectively. According to Khayatkhoei and Elgammal~\cite{khayatkhoei2020spatial}, we can simplify the Eq. ~\ref{eqn:fin_conv} by restricting $\sigma(\cdot)$ to to rectified
linear units (ReLU), which makes the $Conv_{i}^{l} (A^{l})$ become locally piece-wise linear, and absorbing the up-sampling $Up(\cdot)$ into $A_{c}^{l}$. In this way, we transform Eq.~\ref{eqn:fin_conv} to:
\begin{equation}
\label{eqn:fin_sim_conv}
    A_i^{l+1}=Conv_{i}^{l} (A^{l}) = \sum_{c=1}^C F_{ic}^{l} \circledast A_{c}^{l}.
\end{equation}
By applying the 2D discrete Fourier transform (DFT) to $A_i^{l+1}$, it is now viewed in the frequency domain as follows:
\begin{align}
\label{equ:fft_layers}
\tilde{A}_i^{l+1} & =\mathfrak{F}(A_i^{l+1})  = \mathfrak{F}\left(\sum_{c=1}^C F_{ic}^{l} \circledast A_{c}^{l} \right) \nonumber \\
& = \sum_{c=1}^C \mathfrak{F}\left(F_{ic}^{l} \circledast A_{c}^{l} \right) \quad  \text{\emph{(linearity property of FT)}} \nonumber\\
& = \sum_{c=1}^C \mathfrak{F}\left(F_{ic}^{l} \right)  \times \mathfrak{F}\left(A_{c}^{l} \right) \quad  \text{\emph{(conv. property of FT)}} \nonumber \\
& = \sum_{c=1}^C \tilde{F}_{ic}^{l} \times \tilde{A}_{c}^{l}  = \langle \boldsymbol{\tilde{F}_{i}^{l}}, \boldsymbol{\tilde{A}^{l}} \rangle, 
\end{align}
where $\boldsymbol{\tilde{F}_{i}^{l}} = (\tilde{F}_{i1}^{l},...,\tilde{F}_{iC}^{l})^{T}$, and  $\boldsymbol{\tilde{A}^{l}} = (\tilde{A}_{1}^{l},...,\tilde{A}_{C}^{l})^{T}$. Equation~\ref{equ:fft_layers} indicates that in the frequency domain, every channel of in the next layer is decomposed into the  combination all previous layer's channels with different sets of coefficients. 
If we consider ${A}^{l+1}$ as the synthesized output image and fix $\boldsymbol{\tilde{A}^{l}}$ , every vector $\boldsymbol{\tilde{F}^{l}} _{i,i=\overline{1,..C_{(l+1)}}}$ is trained independently to minimized the loss that applied on each $\tilde{A}_i^{l+1}$. When we consider $\boldsymbol{\tilde{A}^{l}}$ as a basic and $\boldsymbol{\tilde{F}_{i}^{l}}$ as the coordinate of $\tilde{A}_i^{l+1}$ with respect to $\boldsymbol{\tilde{A}^{l}}$, to synthesize a new image, the generative models expect that $\boldsymbol{\tilde{A}^{l}}$ is also good enough so that each independent coefficient vector $\boldsymbol{\tilde{F}_{i}^{l}}$ can produce corresponding single channel. In addition, these output channels should become as natural as possible in spacial domain after being stacked together in the order of three color channels: Red, Green and Blue. Small shifts of $\boldsymbol{\tilde{F}^{l}} _{i,i=\overline{1,..C_{(l+1)}}}$ in the frequency domain may only change fine-grained details of visualization but can produce frequency-bias when there is no direct constraint between channels.\\
\subsection{Descriptive Features Extraction}
Let $\mathcal{I}$ be a color image with three channels: Red, Green and Blue, which has width of $W$ and height of $H$. To create its frequency representation, we firstly apply 2D DFT on each channel as follows:
\begin{equation}
\mathfrak{F}_{\mathcal{I}_{R/G/B}}(u,v)=\displaystyle\sum_{x=1}^{W} \sum_{y=1}^{H} \mathcal{I}_{R/G/B} (x, y)\cdot e^{-i2\pi(\frac{ux}{W}+ \frac{vy}{H})},  
\label{eqn:fft}
\end{equation}
where $x$ and $y$, and denote the $x^{th}$ and $y^{th}$ slice in the width and height dimension of $\mathcal{I}$. For convenience, we use the notation $\mathfrak{F}_{\mathcal{I}_{R/G/B}}$ to represent the function that is independently applied for each channel of the image. Note that $\mathfrak{F}_{\mathcal{I}_{R/G/B}}(u,v)$ is now a complex number, \emph{i.e.} $\mathfrak{F}_{\mathcal{I}_{R/G/B}}(u,v) \in \mathbb{C}$, and the spectrum of each channel is obtained as follows:
\begin{equation}
Spec_{R/G/B}(u, v) = mod\left( \mathfrak{F}_{\mathcal{I}_{R/G/B}}(u,v) \right),  
\label{eqn:spec}
\end{equation}
where $mod(\cdot)$ denotes the modulus of complex number.

Although it would be challenging for human eyes to distinguish between real and GAN-generated fake images, we believe that their frequency spectra differences can be possibly exposed, when we stack the three channels' spectra of real vs. fake. Figure \ref{fig:asynchronous_spec} presents our example of images' spectra from VoxCeleb2 dataset \cite{Nagrani19} and Fake Head Talker dataset \cite{jeon2019faketalkerdetect}. In particular, in the real images, we empirically find that the spectra of three color channels are mostly concurrent when stacking together, whereas they become noisy in the fake images, as shown in Fig. \ref{fig:asynchronous_spec}.

Based on this important observation, we propose the following key statistical descriptive features to discriminate the GAN images in the frequency domain: $Mean$, $Max$, $Min$, $iCorr_{RG}$, $iCorr_{RB}$, and $iCorr_{GB}$, where the details are presented below: 
\begin{itemize}
    \item \textbf{$Mean$}. We take the average of the channel-wise spectrum differences:
    \begin{equation}
        Mean =  \frac{d_{RG} + d_{RB} + d_{GB}}{3},
    \label{eqn:mean}
    \end{equation}
    Also, we use $d_{RG}$ that is the average spectrum differences between the spectra of the Red and Green channel in an image as follows:
    \begin{equation}
        d_{RG} = \frac{1}{WH}  \displaystyle\sum_{u=1}^{W} \sum_{v=1}^{H} \big|Spec_R(u,v)-Spec_G(u,v)\big|, 
    \end{equation}
    and $d_{RB}$ and $d_{GB}$ can be similarly defined.
    \item \textbf{$Max$} and \textbf{$Min$}. We take the maximum and minimum values in $\{d_{RG}, d_{RB}$, $d_{GB}\}$.
    \item \textbf{$i_{Corr_{RG}}$}. We calculate the correlation coefficient between $Spec_R$ and $Spec_G$ and transform it to positive range value by adding 1 to its negative values as follows:
    \begin{equation}
        i_{Corr_{RG}} = - \rho (Spec_R, Spec_G) + 1,
    \end{equation}
    where $\rho$ is the Pearson correlation coefficient, and $iCorr_{RB}$ and $iCorr_{GB}$ can be similarly defined.
\end{itemize}
Our end-to-end pipeline of extracting above frequency descriptive features from a given image is visually illustrated in Fig. \ref{fig:overall_arc}.

\subsection{Binary Classifier}
To demonstrate the characteristic-defining ability of the spectrum disagreement, we first employ the simple classifiers to classify real and fake images as below:
\begin{itemize}
    \item \textbf{Gaussian Mixture Model (GMM)}. GMM is a probabilistic model that assumes the distribution of observed sampling data points is composed of a mixture of many Gaussian distributions, particularly, in our case is two distributions of real and fake class. To determine the means and variances of the two Gaussian distributions, Expectation-Maximization (EM) algorithm is used to iteratively estimate these parameters. Our descriptive features populate in the way that the higher spectrum agreement of an image will have the lower descriptive values. Therefore, we can expect that the Gaussian distribution in the mixture model, which has a smaller expectation will represent the real images' distribution, and the other represents the fake images' distribution. By applying EM, we can classify real and fake images in an unsupervised manner in which the labels of a training dataset are not required.
     \item \textbf{Support Vector Machine (SVM)}. SVM is a robust supervised learning method that maximizes the margin of hyperplanes between different classes. The samples that lie along the margins are called the support vectors. In our experiment, we use SVM with the radial basis function (RBF) kernel to train with our six proposed features.
\end{itemize}

\section{Experiment}
\subsection{Datasets}
To examine the effectiveness of our proposed frequency features, we experiment four types of dataset: Fake Head Talker ~\cite{zakharov2019few}, StyleGAN ~\cite{karras2019style}, StarGAN ~\cite{choi2018stargan}, and Adversarial Latent Auto Encoder (ALAE)~\cite{pidhorskyi2020adversarial}. A brief description of each dataset is provided below as well as in Table 1:
\begin{itemize}
    \item \textbf{Fake Head Talker dataset}~~\cite{zakharov2019few}. Fake Head Talker is generated by the few-shot learning system that is pre-trained extensively on a large dataset (meta-learning). Particularly, their approach includes an embedder, a generator, and a discriminator. After training on a large corpus of talking head videos of different faces with adversarial training, their approach can transform facial landmarks from a source frame into realistically-looking personalized photographs with a few photos of a new target person, and further mimic the target.
    \item \textbf{StyleGAN} dataset~~\cite{karras2019style}. StyleGAN is a high-level style controlling approach that governs its generator through adaptive instance normalization (AdaIN) and  Gaussian noise adding in each convolutional layer. Furthermore, by proposing two novel metrics such as perceptual path length and linear separability, the generated images are less entangled and have different factors of variation.
    \item \textbf{StarGAN} dataset~~\cite{choi2018stargan}. StarGAN is a unified model architecture that is able to train on multiple datasets across different domains. By proposing a simple mask vector, the StarGAN is able to flexibly utilize multiple datasets containing different label sets, and achieve competitive results in the facial attribute transfer tasks. This new approach with only a single generator and a discriminator has addressed the scalability and robustness limitations of many previous research.
    \item \textbf{Adversarial Latent Auto Encoder (ALAE)}  dataset ~~\cite{pidhorskyi2020adversarial}. ALAE is an autoencoder-based generative model that is capable to learn the disentangled representations in the latent space with adversarial settings. The ALAE model can not only synthesize high-resolution images comparing with StyleGAN, but also can further manipulate or reconstruct the new input facial images.
\end{itemize}

Details of each dataset in our experiment are summarized in Table ~\ref{table:datasets}. In our experiment, the number of real and GAN fake images are equal in both training and test sets. We further provide the histograms to visualize the distributions of six descriptive features of these datasets in Supplementary material.
\begin{table*}[h]
\caption{Details of datasets used in our experiment.} 
\centering
\begin{tabular}{l c c c c } 
\Xhline{3\arrayrulewidth}
\multirow{2}{*}{Datasets} & \multirow{2}{*}{Resolution} & \multirow{2}{*}{Source datasets} & Training size  & Test size  \\ 
                                                                                     & &  &(real+ fake)   & (real+ fake)\\
\Xhline{2\arrayrulewidth}
Fake Head Talker ~\cite{zakharov2019few}    & $224 \times 224$   & VoxCeleb2 ~\cite{Nagrani19}  & 18,800 & 18,800    \\
StyleGAN ~\cite{karras2019style}     &  $1024 \times 1024$      & FFHQ \footnote{\url{https://github.com/NVlabs/ffhq-dataset}}    & 2,000 & 2,000  \\
StarGAN ~\cite{choi2018stargan}    &  $256 \times 256$       & CelebA ~\cite{liu2015faceattributes}     & 2,000 & 1,998  \\
ALAE ~\cite{pidhorskyi2020adversarial}     &  $1024 \times 1024$      & FFHQ      & 2,000 & 2,000  \\
\Xhline{3\arrayrulewidth}
\end{tabular}
\label{table:datasets}
\end{table*}

\subsection{Experimental Results}
To demonstrate the discriminative power of our proposed features, we perform three different experiments.

\textbf{Binary classification}.
Our experimental results are shown in Table~\ref{table:results_1}. We can observe that both unsupervised and supervised methods are able to produce high performance with our newly introduced frequency features. The accuracy scores of the unsupervised method on Fake Head Talker and ALAE dataset are competitive, compared to the supervised approaches. At the same time, they are still higher than $80\%$ on StyleGAN and StarGAN. Meanwhile, the supervised method's accuracy scores are always higher than $95\%$ on the four datasets. We can conclude that our proposed features based on the asynchronous in the frequency spectrum can effectively capture the characteristics of the GAN-generated images, and provide the foundation for distinguishing fake from real images.

\begin{table*}[t]
\caption{Experimental results of GMM and SVM on four datasets with our discriminative features.} 
\centering
\begin{tabular}{l c c c c c c c c} 
\Xhline{3\arrayrulewidth}
\multirow{2}{*}{Datasets}  & \multicolumn{4}{c }{GMM}  & \multicolumn{4}{c  }{SVM} \\ 
\cmidrule(lr){2-5} \cmidrule(lr){6-9}                                                                                     &  Accuracy & Recall & Precision & F1 & Accuracy & Recall  & Precision & F1\\
\Xhline{2\arrayrulewidth}
Fake Head Talker &	0.996 &	1.000 &	0.991 &	0.996 &	0.9972 &	0.994 &	1.000 &	0.997\\
StyleGAN &	0.849 &	0.762 &	0.915 &	0.831 &	0.951 &	0.938 &	0.963 &	0.950 \\
StarGAN &	0.903 &	0.807 &	1.000 &	0.893 &	0.972 &	0.949 &	0.994 &	0.971 \\
ALAE  &	0.992 &	0.984 &	1.000 &	0.992 &	0.999 &	0.997 &	1.000 &	0.998 \\
\Xhline{3\arrayrulewidth}
\end{tabular}
\label{table:results_1}
\end{table*}

\textbf{Unbalanced training datasets}.
Furthermore, to study the feasibility of training with an unbalanced dataset using our features, we gradually reduce the number of fake images in each training dataset to $25\%, 5\%$, and $1\%$ of the total training data size. After that, we apply the SVM as our learning model.  To demonstrate our approach's effectiveness, we compare our method with FakeTalkerDetect model ~\cite{jeon2019faketalkerdetect}, which deployed a pre-trained AlexNet and Siamese network trained on RGB images. The results are presented in Table~\ref{table:results_2}. We can observe that our method with the hand-crafted features outperforms the AlexNet and FakeTalkerDetect on both balanced and unbalanced datasets. Therefore, we can conclude that our simple yet effective features are capable to characterize the fake image much better in the unbalanced training dataset scenario, as well.

\begin{table}[h]
\caption{Comparison between our approach using proposed descriptive features and AlexNet and FakeTalkerDetect method on Fake Head Talker dataset. The precision, recall and F1 scores of AlexNet and FakeTalkerDetect from~\protect\cite{jeon2019faketalkerdetect}, and their values are rounded to second decimal}
\centering
\resizebox{.48\textwidth}{!}{%
\begin{tabular}{l c c c c } 
\Xhline{3\arrayrulewidth}
Methods & Accuracy & Recall & Precision  & F1\\
\Xhline{2\arrayrulewidth}
AlexNet ($50\%$ fake) &	0.981 &	0.98 &	0.98 &	0.98 \\ 
FakeTalkerDetect &	0.984 &	0.98 &	0.98 &	0.98 \\
SVM (ours) &	\textbf{0.997} &	\textbf{0.994} &	\textbf{1.00} &	\textbf{0.997} \\
\cmidrule(lr){1-5}
AlexNet ($25\%$ fake) &	0.971 &	0.95 &	0.95 &	0.96 \\
FakeTalkerDetect &	0.986 &	0.98 &	0.98 &	0.98 \\
SVM (ours) &	\textbf{0.998} &	\textbf{0.995} &	\textbf{1.000} &	\textbf{0.997} \\
\cmidrule(lr){1-5}
AlexNet ($5\%$ fake) &	0.964 &	0.98 &	0.80 &	0.87 \\
FakeTalkerDetect &	0.988 &	0.99 &	0.91 &	0.94 \\
SVM (ours) &	\textbf{0.997} &	 \textbf{0.997} &	\textbf{0.997} &	\textbf{0.997} \\
\cmidrule(lr){1-5}
AlexNet ($1\%$ fake) &	0.963 &	0.98 &	0.61 &	0.67 \\
FakeTalkerDetect &	0.988 &	0.99 &	0.74 &	0.82 \\
SVM (ours) &	\textbf{0.992} &	\textbf{0.999} &	\textbf{0.986} &	\textbf{0.992} \\
\Xhline{3\arrayrulewidth}
\end{tabular}%
}
\label{table:results_2}
\end{table}

\textbf{Unsupervised domain adaptation}. In this task, we propose an algorithm using our proposed features that allows a pre-trained SVM model on one source dataset (\emph{e.g.,} StyleGAN) can detect fake images in a new target dataset (\emph{e.g.,} ALAE) with the only prior knowledge of the target feature expectations.

In particular, we first take the two Gaussian expectation values of two mixture distributions of each feature from both source and target dataset. These expectation values are kept as our prior knowledge about the target dataset. We then scale the source training set features such that their two Gaussian expectation values are normalized between 0 and 1 , to better fit the training dataset with the SVM model. In the testing phase, with our prior knowledge above, we can scale the testing features from the target dataset using the known expectation values and feed them to the pre-trained SVM model to make prediction. This adaptation learning process is summarized in the Algorithm ~\ref{alg:domain_adatation}. 

\begin{table*}[h]
\caption{Experimental results of domain adaptation task  using our proposed features}
\centering
\begin{tabular}{l l c c c c } 
\Xhline{3\arrayrulewidth}
Source dataset & Target dataset & Accuracy & Recall & Precision  & F1\\
\Xhline{2\arrayrulewidth}
\multirow{3}{*}{Fake Head Talker}
        & StyleGAN &	0.800 &	0.908 &	0.745 &	0.819\\
        & StarGAN &	0.918 &	0.844 &	0.992 &	0.912\\
        & ALAE &	0.994 &	0.995 &	0.993 &	0.994\\
        \cmidrule(lr){1-6}
\multirow{3}{*}{StyleGAN} 
        & Fake Head Talker &	0.965 &	0.932 &	0.998 &	0.964\\
        & StarGAN	 & 0.906 &	0.814 &	0.998 &	0.896\\
        & ALAE &	0.991 &	0.982 &	1.000 &	0.991\\
        \cmidrule(lr){1-6}
\multirow{3}{*}{StarGAN} 
        & Fake Head Talker &	0.983 &	0.982 &	0.983 &	0.983\\
        & StyleGAN &	0.832 &	0.980 &	0.756 &	0.854\\
        & ALAE &	0.996 &	0.997 &	0.994 &	0.996\\
        \cmidrule(lr){1-6}
\multirow{3}{*}{ALAE} 
        & Fake Head Talker &	0.993 &	0.989 &	0.998 &	0.993\\
        & StyleGAN &	0.890 &	0.955 &	0.845 &	0.897\\
        & StarGAN &	0.929 &	0.871 &	0.985 &	0.925\\
\Xhline{3\arrayrulewidth}
\end{tabular}
\label{table:results_3}
\end{table*}

\begin{algorithm}[h]
\caption{Unsupervised domain adaptation with SVM using our proposed descriptive features}
\label{alg:domain_adatation}
\begin{algorithmic}[1]
\Require Labeled source set $\{X^s, Y^s\}$, unlabeled target set $X^t$, where $X^{s}$ and $X^{t}$ includes the six proposed features $[f^{s}_1,.., f^{s}_6]$, respectively. The prior knowledge of Gaussian expectation values: $\left [m^{s}_{i, 0}, m^{s}_{i, 1} \right]_{i= \overline{1,\ldots, 6}}$ and $\left[m^{t}_{i, {0}}, m^{t}_{i, {1}}\right]_{i= \overline{1,\ldots, 6}}$.
  \State Step 1: Scale each feature in $X ^s$ and $X ^t$:\par $\Bar{f^{s}_{i}}=\left(f^{s}_{i}-{m}^{s}_{{i}, {0}}\right)/\left( {m}^{s}_{{i}, {1}} - {m}^{s}_{{i}, {0}}\right)$,\par $\Bar{f^{t}_{i}}=\left(f^{t}_{i}-{m}^{t}_{{i}, {0}}\right)/\left( {m}^{t}_{{i}, {1}} - {m}^{t}_{{i}, {0}}\right)$
  \State Step 2: Fit source set $\left\{[\Bar{f^{s}_1},.., \Bar{f^{s}_6}], Y^s \right\}$ with SVM model.
  \State Step 3: Use pre-trained SVM to predict target set label from $[\Bar{f^{t}_1},.., \Bar{f^{t}_6}]$.
\end{algorithmic}
\end{algorithm}

We experiment with the four fake dataset and present the results in Table ~\ref{table:results_3}. We can observe that with our suggested features the pre-trained SVM shows its strong detection ability in the new target domain, where all the detection performance is above $80\%$ of accuracy for any pair of source and target dataset. This preliminary experiment shows that our proposed features can be utilized in domain adaptation tasks with more complex learning models in the future.

\section{Conclusion}
Although GANs have significantly advanced in the past, we discover that there are some areas that GANs' cannot mimic the real images effectively in the frequency domain. Thus, in this work, we propose a preliminary approach that reveals the asynchronous in frequency domain of the three channels in GAN images. By mining statistical features in frequency domain, our simple yet effective unsupervised and supervised learning methods can easily discriminate the real and GAN-based synthetic facial images without utilizing deep learning methods. Our extensive experiments demonstrates that the proposed features' power in three scenarios: 1) unsupervised and supervised binary classification, 2) unbalanced training dataset, and 3) domain adaptation task. For future work, we plan to explore and exploit more on these aspects of GAN-generated images to combat against misuses from attackers, and extend our work to deepfake detection.

\begin{acknowledgments}
  This work was partly supported by Institute of Information \& communications Technology Planning \& Evaluation (IITP) grant funded by the Korea government (MSIT) (No.2019-0-00421, AI Graduate School Support Program (Sungkyunkwan University)), (No. 2019-0-01343, Regional strategic industry convergence security core talent training business) and the Basic Science Research Program through National Research Foundation of Korea (NRF) grant funded by Korea government MSIT (No. 2020R1C1C1006004). Additionally, this research was partly supported by IITP grant funded by the Korea government MSIT (No. 2021-0-00017, Original Technology Development of Artificial Intelligence Industry) and was partly supported by the Korea government MSIT, under the High-Potential Individuals Global Training Program (2019-0-01579) supervised by the IITP.
\end{acknowledgments}

\bibliography{ceurart}

\end{document}